\documentclass[10pt,twocolumn,letterpaper]{article}

\usepackage{iccv}

\newcommand{\finalcopy}{\toggletrue{iccvfinal}}

\makeatletter
\@namedef{ver@everyshi.sty}{}
\makeatother

\usepackage[dvipsnames,svgnames,x11names]{xcolor}
\usepackage{tikz}
\usetikzlibrary{arrows.meta,shapes,calc,matrix,fit,positioning,backgrounds,decorations.markings}
\usepackage{pgfplots}
\usepackage{pgfplotstable}
\pgfplotsset{compat=1.9}
\usepackage{xstring}

\usepgfplotslibrary{external}

\IfBeginWith*{\jobname}{fig/extern/}{\finalcopy}{}

\tikzstyle{every picture}+=[
	remember picture,
	every text node part/.style={align=center},
	every matrix/.append style={ampersand replacement=\&},
]
\tikzstyle{tight} = [inner sep=0pt,outer sep=0pt]
\tikzstyle{node}  = [draw,circle,tight,minimum size=12pt,anchor=center]
\tikzstyle{op}    = [draw,circle,tight]
\tikzstyle{dot}   = [fill,draw,circle,inner sep=1pt,outer sep=0]
\tikzstyle{pt}    = [fill,draw,circle,inner sep=1.5pt,outer sep=.2pt]
\tikzstyle{box}   = [draw,thick,rectangle,inner sep=3pt]
\tikzstyle{high}  = [black!60]
\tikzstyle{group} = [high,box,opacity=.5]
\tikzstyle{rectc} = [tight,transform shape]
\tikzstyle{rect}  = [rectc,anchor=south west]

\tikzset{every mark/.append style={solid}}
\pgfplotsset{
	grid=both, width=\columnwidth, try min ticks=5,
	every axis/.append style={font=\small},
	every axis plot/.append style={thick,mark=none,mark size=1.8,tension=0.18},
	legend cell align=left, legend style={fill opacity=0.8},
	xticklabel={\pgfmathprintnumber[assume math mode=true]{\tick}},
	yticklabel={\pgfmathprintnumber[assume math mode=true]{\tick}},
	nodes near coords math/.style={
		nodes near coords={\pgfmathprintnumber[assume math mode=true]{\pgfplotspointmeta}},
	},
}

\pgfplotsset{
	dash/.style={mark=o,dashed,opacity=0.6},
	dott/.style={mark=o,dotted,opacity=0.6},
	nolim/.style={enlargelimits=false},
	plain/.style={every axis plot/.append style={},nolim,grid=none},
}

\makeatletter
\renewcommand\paragraph{\@startsection{paragraph}{4}{\z@}{1ex}{-1em}{\normalfont\normalsize\bfseries}}
\makeatother

\usepackage{times}
\usepackage{epsfig}
\usepackage{graphicx}
\usepackage{amsmath}
\usepackage{amssymb}
\usepackage{multirow}
\usepackage{subcaption}
\usepackage{enumitem}
\usepackage{booktabs}
\usepackage{balance}

\usepackage[pagebackref=true,breaklinks=true,letterpaper=true,colorlinks,bookmarks=false]{hyperref}

\iccvfinalcopy 

\ificcvfinal\fi

\begin{document}

\title{On the hidden treasure of dialog in video question answering}

\author{Deniz Engin$^{1,2}$ \qquad Fran{ç}ois Schnitzler$^{2}$ \qquad Ngoc Q. K. Duong$^{2}$ \qquad  Yannis Avrithis$^{1}$ \\ 
$^{1}$Inria, Univ Rennes, CNRS, IRISA \qquad $^{2}$InterDigital}

\maketitle
\ificcvfinal\thispagestyle{empty}\fi


\newcommand{\head}[1]{{\smallskip\noindent\textbf{#1}}}
\newcommand{\alert}[1]{{\color{red}{#1}}}
\newcommand{\sm}{\scriptsize}
\newcommand{\eq}[1]{(\ref{eq:#1})}

\newcommand{\Th}[1]{\textsc{#1}}
\newcommand{\mr}[2]{\multirow{#1}{*}{#2}}
\newcommand{\mc}[2]{\multicolumn{#1}{c}{#2}}
\newcommand{\tb}[1]{\textbf{#1}}
\newcommand{\ch}{\checkmark}

\newcommand{\red}[1]{{\color{red}{#1}}}
\newcommand{\blue}[1]{{\color{blue}{#1}}}
\newcommand{\green}[1]{{\color{green}{#1}}}
\newcommand{\gray}[1]{{\color{gray}{#1}}}

\newcommand{\citeme}[1]{\red{[XX]}}
\newcommand{\refme}[1]{\red{(XX)}}

\newcommand{\fig}[2][1]{\includegraphics[width=#1\linewidth]{fig/#2}}
\newcommand{\figh}[2][1]{\includegraphics[height=#1\linewidth]{fig/#2}}


\newcommand{\tran}{^\top}
\newcommand{\mtran}{^{-\top}}
\newcommand{\zcol}{\mathbf{0}}
\newcommand{\zrow}{\zcol\tran}

\newcommand{\ind}{\mathbbm{1}}
\newcommand{\expect}{\mathbb{E}}
\newcommand{\nat}{\mathbb{N}}
\newcommand{\zahl}{\mathbb{Z}}
\newcommand{\real}{\mathbb{R}}
\newcommand{\proj}{\mathbb{P}}
\newcommand{\prob}{\mathbf{Pr}}
\newcommand{\normal}{\mathcal{N}}

\newcommand{\mif}{\textrm{if}\ }
\newcommand{\other}{\textrm{otherwise}}
\newcommand{\minimize}{\textrm{minimize}\ }
\newcommand{\maximize}{\textrm{maximize}\ }
\newcommand{\st}{\textrm{subject\ to}\ }

\newcommand{\id}{\operatorname{id}}
\newcommand{\const}{\operatorname{const}}
\newcommand{\sgn}{\operatorname{sgn}}
\newcommand{\var}{\operatorname{Var}}
\newcommand{\mean}{\operatorname{mean}}
\newcommand{\trace}{\operatorname{tr}}
\newcommand{\diag}{\operatorname{diag}}
\newcommand{\vect}{\operatorname{vec}}
\newcommand{\cov}{\operatorname{cov}}
\newcommand{\sign}{\operatorname{sign}}
\newcommand{\prj}{\operatorname{proj}}

\newcommand{\softmax}{\operatorname{softmax}}
\newcommand{\clip}{\operatorname{clip}}

\newcommand{\defn}{\mathrel{:=}}
\newcommand{\peq}{\mathrel{+\!=}}
\newcommand{\meq}{\mathrel{-\!=}}

\newcommand{\floor}[1]{\left\lfloor{#1}\right\rfloor}
\newcommand{\ceil}[1]{\left\lceil{#1}\right\rceil}
\newcommand{\inner}[1]{\left\langle{#1}\right\rangle}
\newcommand{\norm}[1]{\left\|{#1}\right\|}
\newcommand{\abs}[1]{\left|{#1}\right|}
\newcommand{\frob}[1]{\norm{#1}_F}
\newcommand{\card}[1]{\left|{#1}\right|\xspace}
\newcommand{\diff}{\mathrm{d}}
\newcommand{\der}[3][]{\frac{d^{#1}#2}{d#3^{#1}}}
\newcommand{\pder}[3][]{\frac{\partial^{#1}{#2}}{\partial{#3^{#1}}}}
\newcommand{\ipder}[3][]{\partial^{#1}{#2}/\partial{#3^{#1}}}
\newcommand{\dder}[3]{\frac{\partial^2{#1}}{\partial{#2}\partial{#3}}}

\newcommand{\wb}[1]{\overline{#1}}
\newcommand{\wt}[1]{\widetilde{#1}}

\def\xssp{\hspace{1pt}}
\def\ssp{\hspace{3pt}}
\def\msp{\hspace{5pt}}
\def\lsp{\hspace{12pt}}

\newcommand{\cA}{\mathcal{A}}
\newcommand{\cB}{\mathcal{B}}
\newcommand{\cC}{\mathcal{C}}
\newcommand{\cD}{\mathcal{D}}
\newcommand{\cE}{\mathcal{E}}
\newcommand{\cF}{\mathcal{F}}
\newcommand{\cG}{\mathcal{G}}
\newcommand{\cH}{\mathcal{H}}
\newcommand{\cI}{\mathcal{I}}
\newcommand{\cJ}{\mathcal{J}}
\newcommand{\cK}{\mathcal{K}}
\newcommand{\cL}{\mathcal{L}}
\newcommand{\cM}{\mathcal{M}}
\newcommand{\cN}{\mathcal{N}}
\newcommand{\cO}{\mathcal{O}}
\newcommand{\cP}{\mathcal{P}}
\newcommand{\cQ}{\mathcal{Q}}
\newcommand{\cR}{\mathcal{R}}
\newcommand{\cS}{\mathcal{S}}
\newcommand{\cT}{\mathcal{T}}
\newcommand{\cU}{\mathcal{U}}
\newcommand{\cV}{\mathcal{V}}
\newcommand{\cW}{\mathcal{W}}
\newcommand{\cX}{\mathcal{X}}
\newcommand{\cY}{\mathcal{Y}}
\newcommand{\cZ}{\mathcal{Z}}

\newcommand{\vA}{\mathbf{A}}
\newcommand{\vB}{\mathbf{B}}
\newcommand{\vC}{\mathbf{C}}
\newcommand{\vD}{\mathbf{D}}
\newcommand{\vE}{\mathbf{E}}
\newcommand{\vF}{\mathbf{F}}
\newcommand{\vG}{\mathbf{G}}
\newcommand{\vH}{\mathbf{H}}
\newcommand{\vI}{\mathbf{I}}
\newcommand{\vJ}{\mathbf{J}}
\newcommand{\vK}{\mathbf{K}}
\newcommand{\vL}{\mathbf{L}}
\newcommand{\vM}{\mathbf{M}}
\newcommand{\vN}{\mathbf{N}}
\newcommand{\vO}{\mathbf{O}}
\newcommand{\vP}{\mathbf{P}}
\newcommand{\vQ}{\mathbf{Q}}
\newcommand{\vR}{\mathbf{R}}
\newcommand{\vS}{\mathbf{S}}
\newcommand{\vT}{\mathbf{T}}
\newcommand{\vU}{\mathbf{U}}
\newcommand{\vV}{\mathbf{V}}
\newcommand{\vW}{\mathbf{W}}
\newcommand{\vX}{\mathbf{X}}
\newcommand{\vY}{\mathbf{Y}}
\newcommand{\vZ}{\mathbf{Z}}

\newcommand{\va}{\mathbf{a}}
\newcommand{\vb}{\mathbf{b}}
\newcommand{\vc}{\mathbf{c}}
\newcommand{\vd}{\mathbf{d}}
\newcommand{\ve}{\mathbf{e}}
\newcommand{\vf}{\mathbf{f}}
\newcommand{\vg}{\mathbf{g}}
\newcommand{\vh}{\mathbf{h}}
\newcommand{\vi}{\mathbf{i}}
\newcommand{\vj}{\mathbf{j}}
\newcommand{\vk}{\mathbf{k}}
\newcommand{\vl}{\mathbf{l}}
\newcommand{\vm}{\mathbf{m}}
\newcommand{\vn}{\mathbf{n}}
\newcommand{\vo}{\mathbf{o}}
\newcommand{\vp}{\mathbf{p}}
\newcommand{\vq}{\mathbf{q}}
\newcommand{\vr}{\mathbf{r}}
\newcommand{\Vs}{\mathbf{s}}
\newcommand{\vt}{\mathbf{t}}
\newcommand{\vu}{\mathbf{u}}
\newcommand{\vv}{\mathbf{v}}
\newcommand{\vw}{\mathbf{w}}
\newcommand{\vx}{\mathbf{x}}
\newcommand{\vy}{\mathbf{y}}
\newcommand{\vz}{\mathbf{z}}

\newcommand{\vone}{\mathbf{1}}
\newcommand{\vzero}{\mathbf{0}}

\newcommand{\valpha}{{\boldsymbol{\alpha}}}
\newcommand{\vbeta}{{\boldsymbol{\beta}}}
\newcommand{\vgamma}{{\boldsymbol{\gamma}}}
\newcommand{\vdelta}{{\boldsymbol{\delta}}}
\newcommand{\vepsilon}{{\boldsymbol{\epsilon}}}
\newcommand{\vzeta}{{\boldsymbol{\zeta}}}
\newcommand{\veta}{{\boldsymbol{\eta}}}
\newcommand{\vtheta}{{\boldsymbol{\theta}}}
\newcommand{\viota}{{\boldsymbol{\iota}}}
\newcommand{\vkappa}{{\boldsymbol{\kappa}}}
\newcommand{\vlambda}{{\boldsymbol{\lambda}}}
\newcommand{\vmu}{{\boldsymbol{\mu}}}
\newcommand{\vnu}{{\boldsymbol{\nu}}}
\newcommand{\vxi}{{\boldsymbol{\xi}}}
\newcommand{\vomikron}{{\boldsymbol{\omikron}}}
\newcommand{\vpi}{{\boldsymbol{\pi}}}
\newcommand{\vrho}{{\boldsymbol{\rho}}}
\newcommand{\vsigma}{{\boldsymbol{\sigma}}}
\newcommand{\vtau}{{\boldsymbol{\tau}}}
\newcommand{\vupsilon}{{\boldsymbol{\upsilon}}}
\newcommand{\vphi}{{\boldsymbol{\phi}}}
\newcommand{\vchi}{{\boldsymbol{\chi}}}
\newcommand{\vpsi}{{\boldsymbol{\psi}}}
\newcommand{\vomega}{{\boldsymbol{\omega}}}

\newcommand{\rLambda}{\mathrm{\Lambda}}
\newcommand{\rSigma}{\mathrm{\Sigma}}

\newcommand{\vLambda}{\bm{\rLambda}}
\newcommand{\vSigma}{\bm{\rSigma}}

\makeatletter
\newcommand*\bdot{\mathpalette\bdot@{.7}}
\newcommand*\bdot@[2]{\mathbin{\vcenter{\hbox{\scalebox{#2}{$\m@th#1\bullet$}}}}}
\makeatother

\makeatletter
\DeclareRobustCommand\onedot{\futurelet\@let@token\@onedot}
\def\@onedot{\ifx\@let@token.\else.\null\fi\xspace}

\def\eg{\emph{e.g}\onedot} \def\Eg{\emph{E.g}\onedot}
\def\ie{\emph{i.e}\onedot} \def\Ie{\emph{I.e}\onedot}
\def\cf{\emph{cf}\onedot} \def\Cf{\emph{Cf}\onedot}
\def\etc{\emph{etc}\onedot} \def\vs{\emph{vs}\onedot}
\def\wrt{w.r.t\onedot} \def\dof{d.o.f\onedot} \def\aka{a.k.a\onedot}
\def\etal{\emph{et al}\onedot}
\makeatother

\def\episodeSum{episode dialog summary\xspace}
\def\episodeSums{episode dialog summaries\xspace}
\def\sceneSum{scene dialog summary\xspace}
\def\sceneSums{scene dialog summaries\xspace}
\def\branch{stream\xspace}
\def\branches{streams\xspace}
\def\up[#1]{\expandafter\MakeUppercase#1}

\newcommand{\tok}{\operatorname{tok}}
\newcommand{\att}{\text{att}}
\newcommand{\best}[1]{{\color{red!60!black}{#1}}}

\begin{abstract}
High-level understanding of stories in video such as movies and TV shows from raw data is extremely challenging. Modern video question answering (VideoQA) systems often use additional human-made sources like plot synopses, scripts, video descriptions or knowledge bases. In this work, we present a new approach to understand the whole story without such external sources. The secret lies in the dialog: unlike any prior work, we treat dialog as a noisy source to be converted into text description via \emph{dialog summarization}, much like recent methods treat video. The input of each modality is encoded by transformers independently, and a simple fusion method combines all modalities, using soft temporal attention for localization over long inputs. Our model outperforms the state of the art on the KnowIT VQA dataset by a large margin, without using question-specific human annotation or human-made plot summaries. It even outperforms human evaluators who have never watched any whole episode before. Code is available at \url{https://engindeniz.github.io/dialogsummary-videoqa}
\end{abstract}

\section{Introduction}
\label{sec:intro}

\begin{figure}
\begin{center}
	\fig{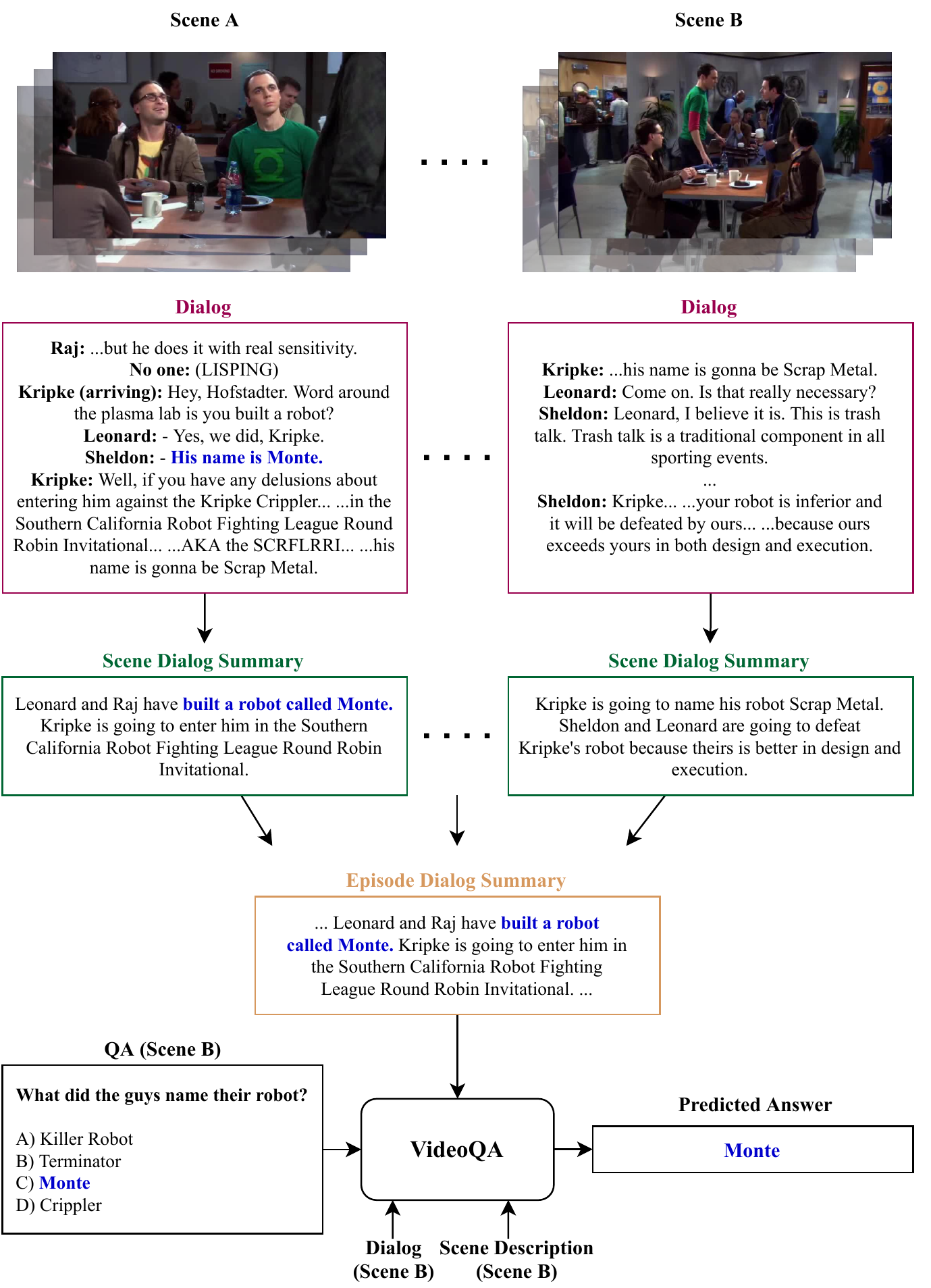}
\end{center}
\caption{In VideoQA, a question is associated with Scene B, but it can only be answered by information from Scene A. We generate \episodeSums from subtitles and give them as input to our VideoQA system, dispensing with the need for external knowledge.}
\label{fig:idea}
\end{figure}

Deep learning has accelerated progress in vision and language tasks. \emph{Visual-semantic embeddings}~\cite{kiros2014unifying,FCS+13} have allowed zero-shot learning, cross-modal retrieval and generating new descriptions from embeddings. \emph{Image captioning}~\cite{vinyals2015show} and \emph{visual question answering }(VQA)~\cite{AAL+15} have demonstrated generation of realistic natural language description of images and a great extent of multimodal semantic understanding. The extension to \emph{video captioning}~\cite{Krishna_2017_ICCV,venugopalan2015sequence} and \emph{video question answering} (VideoQA)~\cite{TZS+16,lei2018tvqa} has enabled further progress because video requires a higher level of reasoning to understand complex events~\cite{zellers2019recognition}.

VideoQA systems typically have similar architecture focusing on multimodal embeddings/description, temporal attention and localization, multimodal fusion and reasoning. While it is often hard to isolate progress in individual components, there are some clear trends. For instance, custom self-attention and memory mechanisms for fusion and reasoning~\cite{Na_2017_ICCV,KCKZ18,FZZ+19} are gradually being streamlined by using \emph{transformer} architectures~\cite{UMDS20,Kim_2020_CVPR,YGC+20}; while visual embeddings~\cite{TZS+16} are being replaced by semantic embeddings~\cite{lei2018tvqa} and \emph{text descriptions} by captioning~\cite{kim2020dense,Chadha2020iPerceive}.

Datasets are essential for progress in the field, but often introduce bias. For instance, questions from text summaries are less relevant to visual information~\cite{TZS+16}; supervised temporal localization~\cite{lei2018tvqa} biases system design towards two-stage localization$\to$answering~\cite{lei2019tvqa+,Kim_2020_CVPR}; fixed question structure focusing on temporal localization~\cite{lei2018tvqa} often results in mere \emph{alignment} of questions with subtitles and \emph{matching} answers with the discovered context~\cite{kim2020dense}, providing little progress on the main objective, which is to study the level of understanding.

Bias can be removed by removing localization supervision and balancing questions over different aspects of comprehension, for instance visual, textual, or semantic~\cite{garcia2020knowit}. However, the requirement of external knowledge, which can be in the form of hints or even ground truth, does not leave much progress in inferring such knowledge from raw data~\cite{garcia2020knowit}. Even weakening this requirement to plain text \emph{human-generated summaries}~\cite{garcia2020knowledge}, still leaves a system unusable in the absence of such data.

In many cases, as illustrated in \autoref{fig:idea}, a question on some part of a story may require knowledge that can be recovered from dialog in other parts of the story. However, despite being textual, raw dialog is often informal and repetitive; searching over all available duration of such noisy source is error-prone and impractical. Inspired by the trend of video captioning, we go a step further and apply the same idea to \emph{dialog}: We \emph{summarize} raw dialog, converting it into \emph{text description} for question answering.

Our finding is astounding: our dialog summary is not only a valid replacement for human-generated summary in handling questions that require knowledge on a whole story, but it outperforms them by a large margin.

Our contributions can be summarized as follows:
\begin{enumerate}[itemsep=2pt, parsep=0pt, topsep=3pt]
    \item We apply \emph{dialog summarization} to video question answering for the first time (\autoref{sec:dialog}).
    \item Building on a modern VideoQA system, we convert all input sources into \emph{plain text description}.
    \item We introduce a weakly-supervised \emph{soft temporal attention} mechanism for localization (\autoref{sec:long}).
    \item We devise a very simple \emph{multimodal fusion} mechanism that has no hyperparameters (\autoref{sec:fusion}).
    \item We set a new state of the art on KnowIT VQA dataset \cite{garcia2020knowit} and we beat non-expert humans for the first time, working only with raw data (\autoref{sec:exp}).
\end{enumerate}

\section{Related Work}
\label{sec:related}


\paragraph{Video Question Answering} 

Progress on video question answering has been facilitated and driven by several datasets and benchmarks. VideoQA by Tapaswi \etal~\cite{TZS+16} addresses answering questions created from \emph{plot synopses} using a variety of input sources, including video, subtitles, scene descriptions, scripts and the plot synopses themselves. Methods experimenting on MovieQA focus on \emph{memory networks} capturing information from the \emph{whole movie} by videos and subtitles~\cite{Na_2017_ICCV,kim2019progressive}, scene-based memory attention networks to learn joint representations of frames and captions~\cite{KCKZ18}, and LSTM-based sequence encoders to learn visual-text embeddings~\cite{liang2018focal}. 

TVQA~\cite{lei2018tvqa} and TVQA+~\cite{lei2019tvqa+} address \emph{scene-based} questions containing \emph{temporal localization} of the answer in TV shows, using video and subtitles. The questions are structured in two parts: one specifying a temporal location in the scene and the other requesting some information from that location. This encourages working with more than one modalities. Methods experimenting on these datasets focus on temporal localization and attention~\cite{lei2019tvqa+,Kim_2020_CVPR}, \emph{captioning}~\cite{kim2020dense,Chadha2020iPerceive} and \emph{transformer}-based pipelines capturing visual-semantic and language information~\cite{YGC+20,UMDS20}.

KnowIT VQA~\cite{garcia2020knowit} is a \emph{knowledge-based} dataset, including questions related to the scene, the episode or the entire story of a TV show, as well as \emph{knowledge annotation} required to address certain questions, in the form of hints. \emph{Transformer}-based methods are proposed to address this task by employing knowledge annotation~\cite{garcia2020knowit} or external human-generated \emph{plot summaries}~\cite{garcia2020knowledge}. Our method differs in substituting human-generated knowledge by summaries automatically generated from raw dialog.


\paragraph{Dialog Summarization}   

Dial2Desc dataset~\cite{pan2018dial2desc} addresses generating \emph{high-level short descriptions from dialog} using a transformer-based text generator. SAMSum corpus~\cite{gliwa2019samsum} is a human-annotated dialog summarization dataset providing speaker information. Methods experimenting on this dataset include existing \emph{document summarization} methods~\cite{gliwa2019samsum}, \emph{graph neural networks} integrating cross-sentence information flow~\cite{zhao2020improving} and graph construction from utterance and commonsense knowledge~\cite{feng2020incorporating}. Since dialog differs from structured text and requires extraction of the conversation structure, recent work focuses on representing the dialog from different \emph{views} by sequence to sequence models~\cite{chen2020multi}. We follow this approach.

\section{Overview}
\label{sec:over}

\begin{figure*}
\centering
    \fig{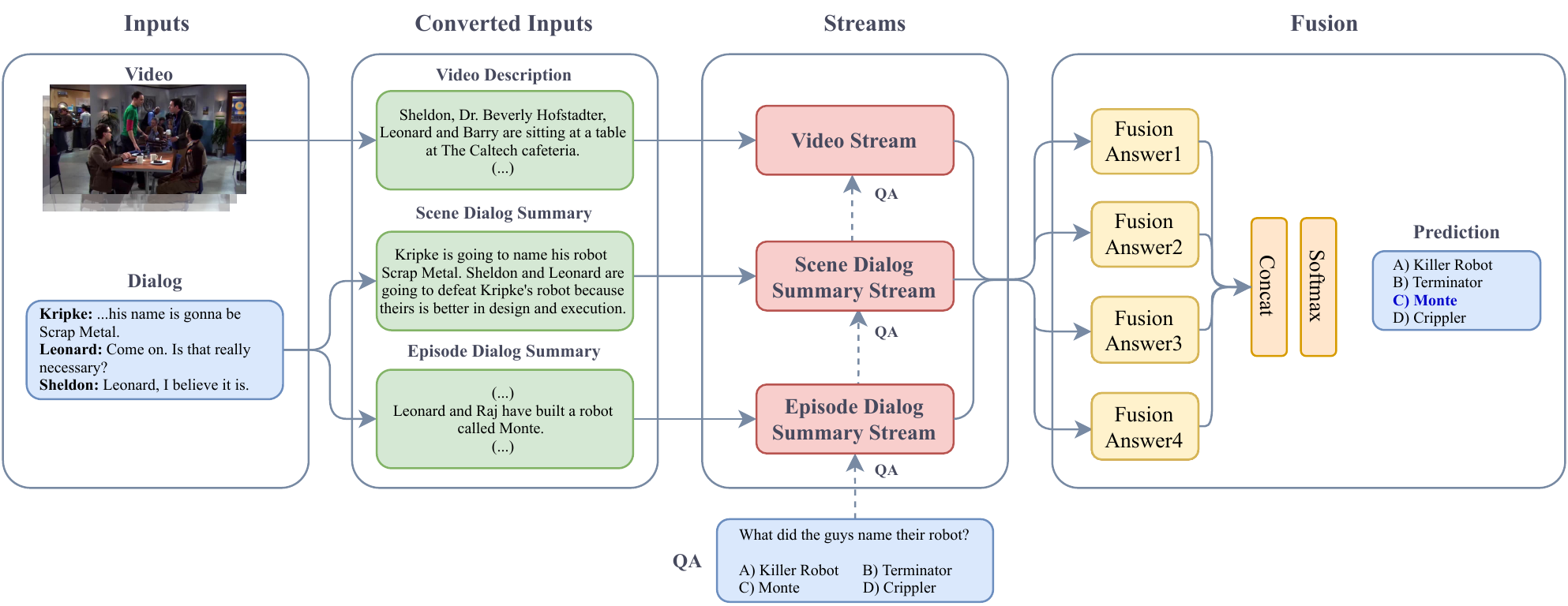}
\caption{Our VideoQA system converts both video and dialog to text descriptions/summaries, the latter at both scene and episode level. Converted inputs are processed independently in \branches, along with the question and each answer, producing a score per answer. Finally, \branch~embeddings are fused separately per answer and a prediction is made.}
\label{fig:over}
\end{figure*}

We address knowledge-based video question answering on TV shows. Each episode is split in \emph{scenes}. For each scene, we are given the \emph{video} (frames) and \emph{dialog} (speaker names followed by subtitle text) and a number of \emph{multiple-choice questions}. Certain questions require high-level understanding of the whole episode or show. Garcia \etal~\cite{garcia2020knowledge}  rely on human-generated \emph{plot summaries} (or \emph{plot} for short), which we use only for comparison. Our objective is to extract the required knowledge from raw data.

As shown in Figure~\ref{fig:over}, we first convert inputs into \emph{plain text description}, including both video (by visual recognition) and dialog (by summarization) (\autoref{sec:input}). A number of separate \emph{\branches} then map text to embeddings, at the level of both \emph{scene} (video and \sceneSum) and \emph{episode} (\episodeSum and plot). The question and answers are embedded together with the input text of each \branch. A \emph{temporal attention} mechanism localizes relevant intervals from episode inputs. Finally, question answering is addressed both in a \emph{single-stream} (\autoref{sec:stream}) and a \emph{multi-stream} (\autoref{sec:fusion}) scenario. The latter amounts to \emph{multi-modal fusion}. We begin our discussion with \emph{transformer} networks (\autoref{sec:trans}), which we use both for dialog summarization and text embeddings in general.
\section{Transformers}
\label{sec:trans}

The \emph{transformer}~\cite{transformervaswani2017attention} is a network architecture that allows for efficient pairwise interaction between input elements. Its main component is an \emph{attention} function, which acts as a form of associative memory.

\emph{Multi-head attention} is a fusion of several attention functions.
The architecture is a stack of multi-head attention, element-wise fully-connected and normalization layers with residual connections. Originally developed for machine translation, it includes an \emph{encoder} and a \emph{decoder} stack. The decoder additionally attends over the output of the encoder stack and is \emph{auto-regressive}, consuming previously generated symbols when generating the next.

BERT~\cite{devlin-etal-2019-bert} is a transformer bidirectional \emph{encoder} only, mapping a sequence of tokens to a sequence of $d$-dimensional vectors. It is pre-trained on unsupervised tasks including prediction of masked tokens and next sentence, and can be also fine-tuned on supervised downstream tasks. It can take a number of \emph{sentences} as in input, where a sentence is an arbitrary span of contiguous text.

We use BERT as the backbone of our model architecture to represent text, using two sentences at a time. Given strings $A$ and $B$, the input is given as
\begin{equation}
	\tok_k(\mathtt{[CLS]} + A + \mathtt{[SEP]} + B + \mathtt{[SEP]}),
\label{eq:bert-in}
\end{equation}
where $+$ is string concatenation and $\tok_k$ is tokenization into $k$ tokens, with zero padding if the input length is less than $k$ and truncation if it is greater. Tokens are represented by WordPiece embeddings~\cite{ScNa12,wu2016google}, concatenated with \emph{position embeddings} representing their position in the input sequence and \emph{segment embeddings}, where segments correspond to sentences and are defined according to occurrences of the \emph{separator} token $\mathtt{[SEP]}$. The output vector in $\real^d$ corresponding to token $\mathtt{[CLS]}$ is an \emph{aggregated representation} of the entire input sequence and we denote it as
\begin{equation}
	f(A, B).
\label{eq:bert-out}
\end{equation}

Sentence-BERT~\cite{reimers-gurevych-2019-sentence} takes a single sentence as input and is trained by \emph{metric learning} objectives, \eg in a siamese or triplet structure, facilitating efficient sentence similarity search. It is learned by fine-tuning a pre-trained BERT model on supervised semantic textual similarity.

BART~\cite{lewis2020bart} combines a bidirectional \emph{encoder} and an auto-regressive \emph{decoder}. It is pre-trained as an unsupervised denoising autoencoder, \ie, corrupting input text and learning to reconstruct the original, and fine-tuned on supervised classification, generation or translation tasks. It is particularly effective on \emph{text generation}, including abstractive dialog, question answering and summarization tasks.

Following~\cite{chen2020multi}, we use sentence-BERT and BART to \emph{segment} and \emph{summarize dialog} respectively.
\section{Input description}
\label{sec:input}

All input sources, \ie, \emph{video}, \emph{dialog} and \emph{plot}, are converted into \emph{plain text description} before being used for question answering. Video is first converted into a \emph{scene graph} by a visual recognition pipeline and then to text description by a set of rules. Importantly, although already in textual form, dialog is also converted into text description by \emph{dialog summarization}. The plot, already in text description form, is used as is, but for comparison only: Our main contribution is to replace human-generated plots by automatically generated descriptions.


\subsection{Dialog}
\label{sec:dialog}

As the main form of human communication, dialog is an essential input source for video understanding and question answering. We use dialog in three ways: \emph{raw dialog} per scene, \emph{dialog summary} per scene and the collection of dialog summary over a whole \emph{episode}.


\paragraph{Raw scene dialog}

As in all prior work, we use the raw dialog associated to the scene of the question, \emph{as is}. Although in textual form, it is \emph{not} a text description. It may still contain more information than dialog summary, which is important to investigate.


\paragraph{\up[\sceneSum]}

Given the dialog associated to the scene of the question, we convert this input source into text description by \emph{dialog summarization}. Despite being of textual form, dialog is very different from text \emph{description}: conversations are often informal, verbose and repetitive, with few utterances being informative; while a description is a narrative in \emph{third-person} point of view with clear information flow structured in paragraphs~\cite{chen2020multi}. Identifying the speaking person is also substantial, especially with multiple people in a conversation. Rather than generic document summarization~\cite{gliwa2019samsum}, we follow a dedicated dialog summarization method~\cite{chen2020multi}, which blends character names with events in the generated summaries.

A dialog is a sequence of \emph{utterances}, each including a \emph{speaker} (character) name and a \emph{sentence} (sequence of tokens). Each utterance is mapped to a vector embedding by Sentence-BERT~\cite{reimers-gurevych-2019-sentence}. The sequence of embeddings over the entire dialog is segmented according to  \emph{topic}, \eg \emph{greetings}, \emph{today's plan}, \etc. by C99~\cite{C99-choi-2000-advances}, as well as \emph{stage}, \eg \emph{opening}, \emph{intention}, \emph{discussion}, \emph{conclusion} by a \emph{hidden Markov model} (HMM)~\cite{althoff2016large}. As a result, for each \emph{view} (topic or stage), the dialog is represented by a sequence of \emph{blocks}, each containing several utterances.

Given the above structure, the input is re-embedded and the summary is generated using an extension of BART~\cite{lewis2020bart}. In particular, there is one \emph{encoder} per view, mapping each block to an embedding. An LSTM~\cite{hochreiter1997long} follows, aggregating the entire view into one embedding, obtained as its last hidden state. The \emph{decoder} attends over the output of each encoder using a \emph{multi-view attention} layer to weight the contribution of each view. It is \emph{auto-regressive}, using previous tokens from ground truth at training and previously predicted tokens by the encoder at inference.

We train the HMM on the dialog sources of our video QA training set; otherwise, we use Sentence-BERT and BART as used/trained by~\cite{chen2020multi}. Once a \sceneSum is generated, it is re-embedded by BERT~\cite{devlin-etal-2019-bert} like all other input sources, as discussed in \autoref{sec:stream}.


\paragraph{\up[\episodeSum]}

We collect the \sceneSums for all scenes of an episode and we concatenate them into an \emph{\episodeSum}. Assuming that the episode of the scene of the question is known, we make available the associated \episodeSum for question answering. This is a long input source and requires \emph{temporal attention}, as discussed in \autoref{sec:long}. Importantly, \episodeSum is our most important contribution in substituting plot summary by an automatically generated description.


\subsection{Plot summary}
\label{sec:plot}
As part of our comparison to~\cite{garcia2020knowledge}, we use publicly available plot summaries\footnote{\url{https://the-big-bang-theory.com/}}, already in text description form. Assuming that the episode of the scene of the question is known, we make available the associated plot \emph{as is}, to help answering \emph{knowledge-based questions}. A plot is shorter and higher-level than our \episodeSum, but it is still long enough to require \emph{temporal attention}. It is important to investigate whether we can dispense of such a human-generated input and how much more information it contains relative to what we can extract automatically. 


\subsection{Video}
\label{sec:video}

We use a visual recognition pipeline to convert raw input video into text description. Following~\cite{garcia2020knowledge}, this pipeline comprises four components: \emph{character recognition}~\cite{schroff2015facenet}, \emph{place recognition}~\cite{zhou2017places}, \emph{object relation detection}~\cite{zhang2019large}, and \emph{action recognition}~\cite{wu2019long}. The outputs of these components are character, place, object, relation and action \emph{nodes}. A directed \emph{video scene graph} is generated by collecting all nodes along with edges and then a textual \emph{scene description} is obtained according to a set of predefined rules.

\section{Single-stream QA}
\label{sec:stream}

As shown in \autoref{fig:over}, there is one \branch per input source, using a transformer to map inputs to embeddings. Following~\cite{garcia2020knowledge}, we first attempt question answering on each \branch alone. In doing so, we learn a linear classifier while fine-tuning the entire transformer representation per \branch. Unlike most existing works, this allows adapting to the data at hand, for instance a particular TV show.

We differentiate \emph{scene} from \emph{episode} inputs, as discussed below. In both cases, the given question and candidate answer strings are denoted as $q$ and $a^c$ for $c=1,\dots,n_c$ respectively, where $n_c$ is the number of candidate answers.


\subsection{Scene input sources}
\label{sec:short}

Scene input sources refer to the scene of the question, \ie, \emph{raw scene dialog}, \emph{\sceneSum} or \emph{video}. The input string is denoted by $x$. For each $c=1,\dots,n_c$, we embed $x$, $q$ and $a^c$ jointly to $d$-dimensional vector
\begin{equation} 
	y^c \defn f(x + q, a^c),
\label{eq:short-bert}
\end{equation}
where $+$ is string concatenation and $f$ is BERT~\eq{bert-out}. A linear classifier with parameters $\vw \in \real^d$, $b \in \real$ yields a score per candidate answer
\begin{equation}
	z^c \defn \vw^\top \cdot y^c + b.
\label{eq:short-class}
\end{equation}
The \emph{score vector} $z \defn (z^1, ..., z^{n_c})$ is followed by softmax and cross-entropy loss. At training, we use $f$ as  pre-trained and we fine-tune it while optimizing $W,b$ on the correct answers of the QA training set. At inference, we predict $\arg\max_c z^c$.


\subsection{Episode input sources}
\label{sec:long}

Episode input sources refer to the entire episode of the scene of the question, \ie, \emph{\episodeSum} and \emph{plot}. Because such input is typically longer than the transformer's maximum sequence length $k$~\eq{bert-in}, we split it into overlapping parts in a \emph{sliding window} fashion. Each part contains the question and one answer, so the window length is $w = k - |q| - |a^c|$. Given an input of length $\ell$ tokens, the number of parts is $n \defn \ceil{\frac{\ell - w}{s}} + 1$, where $s$ is the \emph{stride}. Because all inputs in a mini-batch must have the same number of parts $n_p$ to be stacked in a tensor, certain parts are zero-padded if $n < n_p$ and discarded if $n > n_p$.


\paragraph{Embedding}

The input strings of the parts are denoted by $p_j$ for $j=1,\dots,n_p$. Each part $p_j$ is combined with each candidate answer $a^c$ separately, yielding the $d$-dimensional vectors
\begin{equation} 
	y_j^c \defn f(p_j + q, a^c)
\label{eq:long-bert}
\end{equation}
for $c=1,\dots,n_c$ and $j=1,\dots,n_p$.  A classifier with parameters $\vw \in \real^d$, $b \in \real$ yields a score per candidate answer $c$ and part $j$:
\begin{equation}
	z_j^c \defn \vw\tran \cdot y_j^c + b.
\label{eq:long-class}
\end{equation}


\paragraph{Temporal attention}

At this point, unlike scene inputs~\eq{short-class}, predictions from~\eq{long-class} are not meaningful unless a part $j$ is known, which amounts to \emph{temporal localization} of the part of the input sequence that contains the information needed to answer a question. In TVQA~\cite{lei2018tvqa} and related work~\cite{lei2019tvqa+,kim2020dense,Kim_2020_CVPR}, localization ground truth is available, allowing a two-stage localize-then-answer approach. Without such information, the problem is \emph{weakly supervised}.

Previous work~\cite{garcia2020knowledge} simply chooses the part $j$ corresponding to the maximum score $z_j^c$ over all answers $c$ and all parts $j$ in~\eq{long-class}, which is called \emph{hard temporal attention} in the following. Such hard decision may be harmful when the chosen $j$ is incorrect, especially when the predicted answer happens to be correct, because then the model may receive arbitrary gradient signals at training. To alleviate this, we follow a \emph{soft temporal attention} approach.

In particular, let $S$ be the $n_p \times n_c$ matrix with elements $z_j^c$ over all answers $c$ and all parts $j$~\eq{long-class}. For each part $j$, we take the maximum score over answers
\begin{equation}
    s_j \defn \max_c z_j^c,
\end{equation}
giving rise to a vector $s \defn (s_1,\dots,s_{n_p})$, containing a single best score per part. Then, by soft assignment over the rows of $S$---corresponding to parts---we obtain a score for each answer $c$, represented by \emph{score vector} $z \in \real^c$:
\begin{equation}
    z \defn \softmax(s / T)\tran \cdot S,
\label{eq:temporal-attn}
\end{equation}
where $T$ is a temperature parameter. With this definition of $z$, we have a single score vector and we proceed as in~\eq{short-class}.

\section{Multi-\branch QA}
\label{sec:fusion}

Once a separate transformer has been fine-tuned separately for each \branch, we combine all \branches into a single question answering classifier, which amounts to multi-modal fusion. Here, we introduce two new simple solutions.

In both cases, we freeze all transformers and obtain $d$-dimensional embeddings $y^c$ for each candidate answer $c$ and for each \branch. For scene inputs, $y^c$ is obtained directly from~\eq{short-bert}. Episode input \branches produce $n_p$ embeddings per answer. Temporal localization is thus required for part selection, similar to single stream training. Again, \emph{hard temporal attention} amounts to choosing the part with the highest score according to~\eq{long-class}: $y^c \defn y_{j^*}^c$ where $j^* \defn \arg\max_j(z_j^c)$ and $y_j^c$ is given by~\eq{long-bert}. Instead, similar to~\eq{temporal-attn}, we follow \emph{soft temporal attention}:
\begin{equation}
    y^c \defn \softmax(s / T)\tran \cdot Y_c^{emb},
\label{eq:temporal-attn-multi}
\end{equation}
where $Y_c^{emb}$ is a $n_p \times d$ matrix collecting the embeddings $y_j^c$~\eq{long-bert} of all parts $j$. Finally, for each answer $c$, the embeddings $y^c$ of all \branches are stacked into a $n_s \times d$ \emph{embedding matrix} $Y_c$, where $n_s$ is the number of \branches.
 

\paragraph{Multi-\branch attention}

The columns of $Y_c$ are embeddings of different \branches. We weight them according to weights $w_c \in \real^{n_s}$ obtained from $Y_c$ itself, using a \emph{multi-\branch attention} block, consisting of two fully connected layers followed by softmax:
\begin{equation}
	Y_c^{\att} = \diag(w_c) \cdot Y_c.
\label{eq:fuse-multi}
\end{equation}
For each answer $c$, a fully connected layer maps the $d \times n_s$ matrix $Y_c^{\att}$ to a scalar score. All $n_c$ scores are followed by softmax and cross-entropy loss, whereby the parameters of all layers are jointly optimized.


\paragraph{Self-attention} 

Alternatively, $Y_c$ is mapped to $Y_c^{\att} \in \real^{d \times n_s}$ by a single \emph{multi-head self-attention} block, as in transformers \cite{transformervaswani2017attention}:
\begin{equation}
	Y_c^{\att} = \operatorname{MultiHeadAttention}(Y_c,Y_c,Y_c).
\label{eq:fuse-self}
\end{equation}
The remaining pipeline is the same as in the previous case.


\section{Experiments}
\label{sec:exp}

\subsection{Experimental setup}

\paragraph{Datasets}

The KnowIT VQA~\cite{garcia2020knowit} dataset contains 24,282 human-generated questions associated to 12,087 scenes, each of duration 20 seconds, from 207 episodes of \emph{The Big Bang Theory} TV show. Questions are of four types: \emph{visual} (22\%), \emph{textual} (12\%), \emph{temporal} (4\%) and \emph{knowledge} (62\%). Question types are only known for the test set. Knowledge questions require reasoning based on knowledge from the episode or the entire TV show, which differs from other video question answering datasets. Questions are multiple-choice with $n_c = 4$ answers per question and performance is measured by \emph{accuracy}, per question type and overall. 

\paragraph{Implementation details}

For scene dialog summary generation, we set the minimum sequence length to 30 tokens and the maximum to 100 in the BART~\cite{lewis2020bart} model. With this setting, episode dialog summaries are 2078 tokens long on average, while plot summaries are 659 tokens long.

We fine-tune the BERT$_{\text{BASE}}$~\cite{devlin-etal-2019-bert} uncased model with $N = 12$ transformer blocks, $h = 12$ self-attention heads and embedding dimension $d = 768$ for single-\branch models. The maximum token length $k$ is $512$ for scene, $200$ for plot and $300$ for episode dialog summary inputs. The stride $s$ is $100$ for plot and $200$ for episode dialog summary. The maximum number of parts $n_p$ is $10$ for both. The batch size is $8$ for all single-\branch models and $32$ for multi-\branch. We use SGD with momentum $0.9$ scheduled with initial learning rate $10^{-4}$ for multi-\branch fusions. We use $h = 1$ attention head, and $N = 2$ stacks for self-attention and multi-stream self-attention methods. The number of \branches $n_s$ varies per experiment.

\begin{table*}
\centering
\small
\begin{tabular}{llcccccc} \toprule
	\Th{Method}                                      & \Th{Knowledge}      & \Th{Vis.}  & \Th{Text.} & \Th{Temp.} & \Th{Know.} & \best{\Th{All}}   \\ \midrule
	Rookies~\cite{garcia2020knowit}                  & --                  & 0.936      & 0.932      & 0.624      & 0.655      & 0.748             \\
	Masters~\cite{garcia2020knowit}                  & \ch                 & 0.961      & 0.936      & 0.857      & 0.867      & 0.896             \\ \midrule
	ROCK$_{\text{GT}}$~\cite{garcia2020knowit}       & question GT & 0.747 & \tb{0.819} & 0.756 & 0.708      & 0.731             \\
	ROLL$_{\text{human}}$~\cite{garcia2020knowledge} & question GT & 0.708      & 0.754      & 0.570      & 0.567      & 0.620             \\ \midrule
	TVQA~\cite{lei2018tvqa}                          & --                  & 0.612      & 0.645      & 0.547      & 0.466      & 0.522             \\
	ROCK$_{\text{facial}}$~\cite{garcia2020knowit}   & dataset GT  & 0.654      & 0.688      & 0.628      & 0.646      & 0.652             \\
	ROLL~\cite{garcia2020knowledge}                  & plot                & 0.718      & 0.739      & 0.640      & 0.713      & 0.715             \\ \midrule
	Ours                                             & --                  & \tb{0.755}     & 0.783      & \tb{0.779}     & \tb{0.789}      & \best{\tb{0.781}}             \\
	Ours$_{\text{plot}}$                             & plot                & 0.749      & 0.783  & 0.721      & 0.783 & 0.773 \\ \bottomrule
\end{tabular}
\caption{\emph{State-of-the-art accuracy} on KnowIT VQA. Ours uses the video and \sceneSum as well as the \episodeSum that we generate from the dialog of the entire episode. Ours$_{\text{plot}}$ also uses human-generated plot summaries, like~\cite{garcia2020knowledge}. TVQA uses an LSTM based encoder; all other methods use BERT. Rookies and Masters are humans.}
\label{tab:sota}
\end{table*}

\subsection{Quantitative results}

\autoref{tab:sota} compares of our method with the state of the art. Rookies and Masters are human evaluators: Masters have watched most of the show, whereas Rookies have never watched an episode before~\cite{garcia2020knowit}. TVQA~\cite{lei2018tvqa} encodes visual features and subtitles without considering knowledge information; its results are as reported in~\cite{garcia2020knowit}. ROCK~\cite{garcia2020knowit} uses four visual representations (image, concepts, facial, caption); ROCK$_{\text{facial}}$ is one of its best results. ROCK$_{\text{GT}}$~\cite{garcia2020knowit} and ROLL$_{\text{human}}$~\cite{garcia2020knowledge} use the human knowledge annotation provided by the dataset~\cite{garcia2020knowit}, while ROLL~\cite{garcia2020knowledge} uses human-written plot summaries instead. Our method uses scene video and \sceneSum as well as the \episodeSum that it automatically generates, without any human annotation. Ours$_{\text{plot}}$ additionally uses the same plot as~\cite{garcia2020knowledge}. TVQA uses LSTM; all other methods are based on BERT.

Our method outperforms the best state of the art method (ROLL~\cite{garcia2020knowledge}) by 6.6\%, without any human annotation. By using additional human-generated plots, the gain decreases to 5.8\%. This indicates that our \episodeSum captures the required knowledge and removes the requirement of human-generated input; in fact, human-generated input is harmful. On temporal and knowledge questions in particular, we gain 13.9\% and 7.6\%, respectively, without any human annotation. This implies that our automatically generated \episodeSum increases the understanding of the episode and helps answering all types of questions. Despite ROLL$_{\text{human}}$~\cite{garcia2020knowledge} and ROCK$_{\text{GT}}$~\cite{garcia2020knowit} using ground-truth knowledge, we outperform them by 16.1\%  and 5.0\%, respectively, without any human annotation. We also outperform Rookies, presumably by having access to the dialog of the entire episode. Comparing to Masters, there is still room for improvement.


\subsection{Qualitative analysis}

\begin{figure*}
\centering
\small
\begin{tabular}{cc}
	\fig[.48]{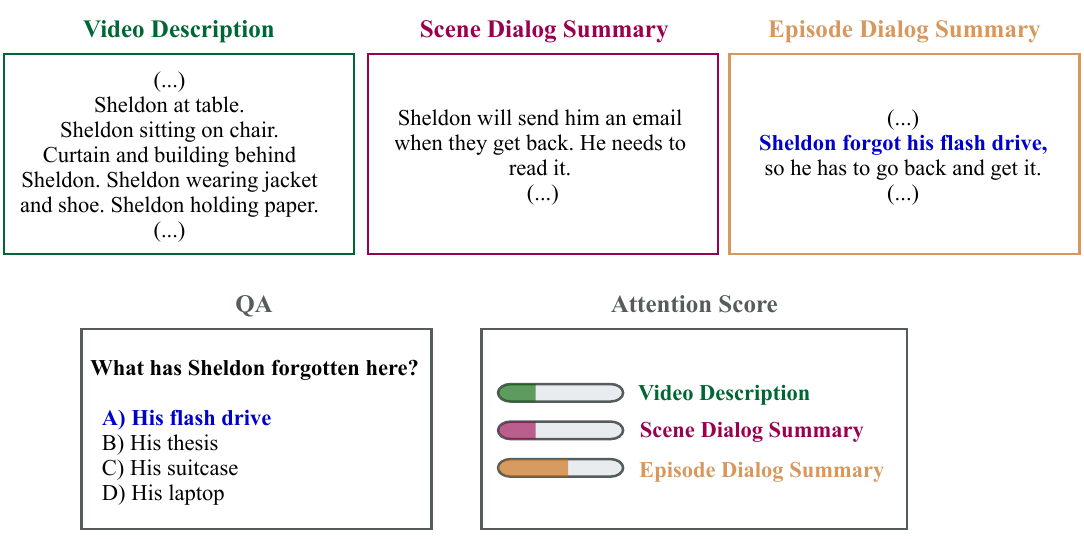} &
	\fig[.48]{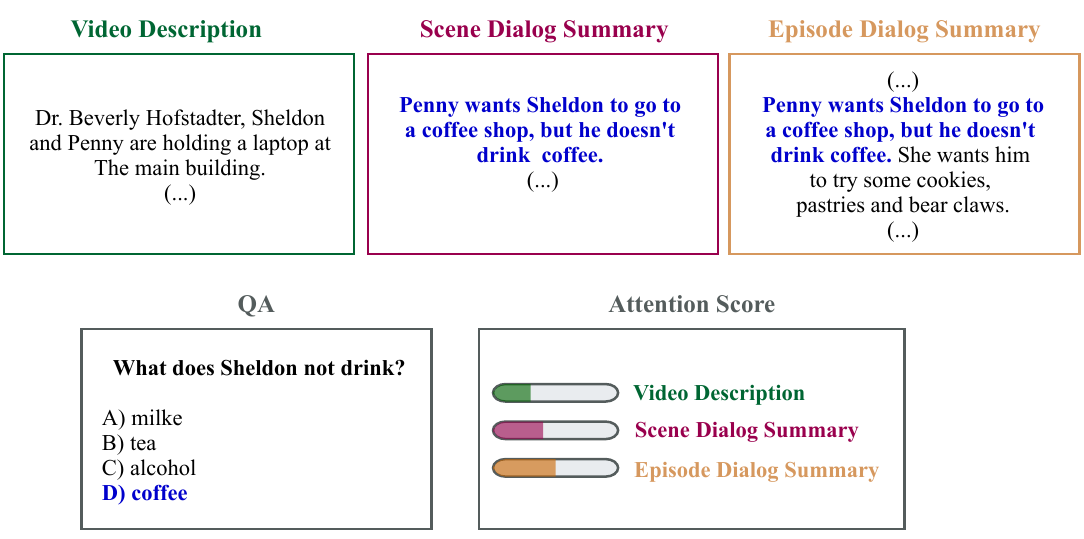} \\
	(a) Knowledge QA &
	(b) Textual QA \\
	\fig[.48]{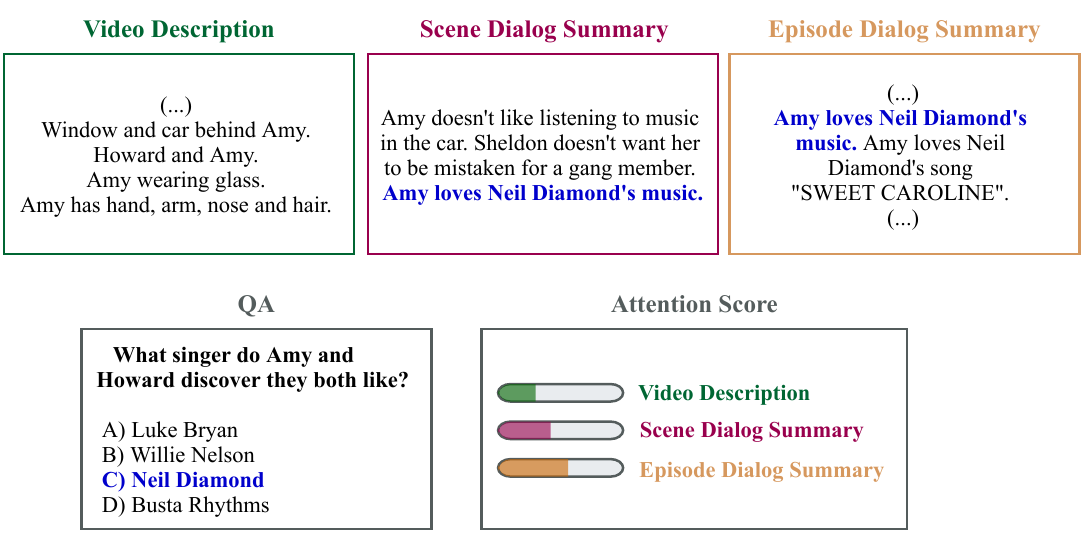} &
	\fig[.48]{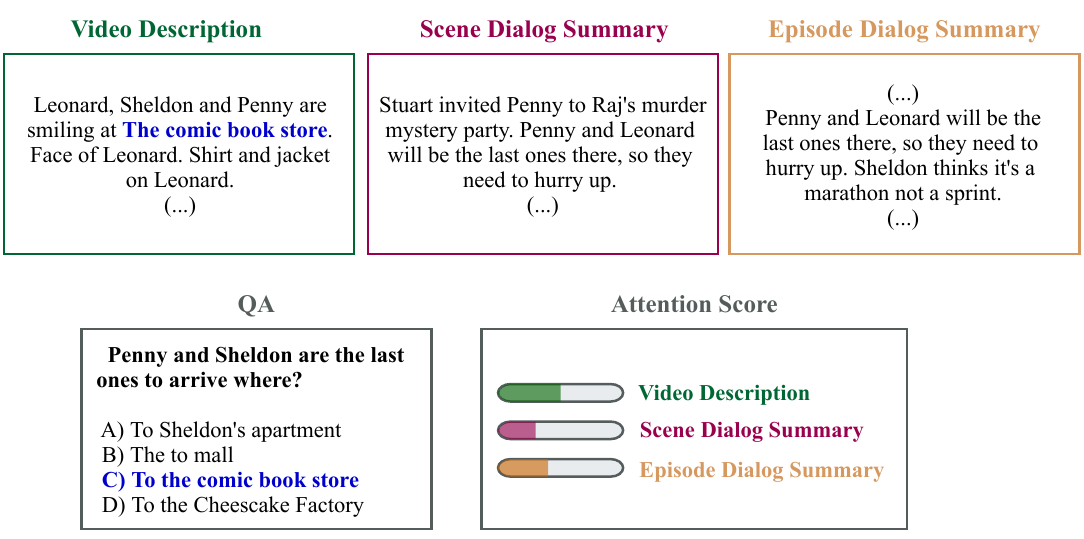} \\
	(c) Temporal QA &
	(d) Visual QA
\end{tabular}
\caption{
\emph{Multi-\branch attention visualization}. We highlight in blue the part of the source text that is relevant to answering the question. The most attended stream is \episodeSum for (a), (b), (c) and video description for (d).}
\label{fig:qual}
\end{figure*}

\autoref{fig:qual} visualizes the correct predictions of our method with \branch attention scores for different question types. In all examples, the model receives three input sources, question/answers and attention scores over inputs. \autoref{fig:qual}(a) shows a \emph{knowledge} question, answered based on \episodeSum, which has the highest attention score. As shown in \autoref{fig:qual}(b), a \emph{textual} question can be answered by using \sceneSum, but also by \episodeSum, since the latter includes the former. \emph{Temporal} questions can be answered from scene inputs such as \sceneSum or video description. According to attention scores, the question in \autoref{fig:qual}(c) is answered by \episodeSum, which includes the correct answer. Finally, \autoref{fig:qual}(d) shows a \emph{visual} question answered by video description.

\begin{table}
\centering
\small
\setlength{\tabcolsep}{4pt}
\begin{tabular}{lcccccc} \toprule
\Th{Method}                                      & \Th{Input}  & \Th{Vis.}  & \Th{Text.} & \Th{Temp.} & \Th{Know.} & \best{\Th{All}}   \\ \midrule
\mr{3}{ROLL~\cite{garcia2020knowledge}}          & D           & 0.656      & 0.772      & 0.570      & 0.525      & 0.584             \\
                                                 & V           & 0.629      & 0.424      & 0.558      & 0.514      & 0.530             \\
                                                 & P           & 0.624      & 0.620      & 0.570      & 0.725      & 0.685             \\ \midrule
\mr{3}{ROLL~\cite{garcia2020knowledge}$\dagger$} & D           & 0.649      & \tb{0.801} & 0.581      & 0.543      & 0.598             \\
                                                 & V           & 0.625      & 0.431      & 0.512      & 0.541      & 0.546             \\
                                                 & P           & 0.647      & 0.554      & 0.674      & 0.694      & 0.667             \\ \midrule
\mr{3}{Ours}                                     & P           & 0.666      & 0.623      & 0.593      & 0.735      & 0.702             \\
                                                 & S           & 0.631      & 0.746      & 0.605      & 0.537      & 0.585             \\
                                                 & E           & \tb{0.676} & 0.750      & \tb{0.779} & \tb{0.785} & \best{\tb{0.756}} \\ \bottomrule
\end{tabular}%
\caption{\emph{Single-\branch QA accuracy} on KnowIT VQA. ROLL~\cite{garcia2020knowledge}: as reported; \cite{garcia2020knowledge}$\dagger$: our reproduction. Our model incorporates the scene dialog and video \branches of the latter as well as the plot, \sceneSum and \episodeSum \branches. Plot differs between \cite{garcia2020knowledge}$\dagger$ and our model by our temporal attention and other improvements (\autoref{tab:roll-improve}). D: dialog; V: video; P: plot; S: \sceneSum; E: \episodeSum. }
\label{tab:branch_results}
\end{table}


\subsection{Ablation studies}
\label{sec:ablation}


\paragraph{Single-\branch results}

\autoref{tab:branch_results} shows our single-\branch QA results. We reproduce~\cite{garcia2020knowledge} for dialog, video, and plot inputs. We replace the plot \branch by one using our new temporal attention (\autoref{sec:long}) and other improvements (\autoref{tab:roll-improve}) and we add two new sources automatically generated from dialog: \sceneSum and \episodeSum. Due to the dataset having a majority of knowledge questions, \episodeSum and plot inputs have higher accuracy than other input sources since they span an entire episode. Our \episodeSum helps in answering questions better than the plot~\cite{garcia2020knowledge}, bringing an accuracy improvement of 5.4\%.

\begin{table}
\centering
\small
\setlength{\tabcolsep}{3pt}
\begin{tabular}{llllll} \toprule
\Th{Method}                  & \Th{Vis.}  & \Th{Text.} & \Th{Temp.} & \Th{Know.} & \best{\Th{All}}   \\ \midrule
Product                      & 0.743      & 0.659      & 0.756      & 0.751      & 0.739             \\
Modality weighting~\cite{garcia2020knowledge}   & 0.708             & \tb{0.786}              & 0.767              & 0.787              & 0.769  \\ \midrule
Self-attention               & \tb{0.759}      & 0.764      & 0.767      & 0.777      & 0.771             \\
Multi-\branch attention      & 0.755      &  0.783      & \tb{0.779} & \tb{0.789} & \best{\tb{0.781}} \\
Multi-\branch self-attn.     & 0.755      & 0.768      & 0.756      & 0.777      & 0.770             \\
\bottomrule
\end{tabular}%
\caption{\emph{Multi-\branch QA accuracy} on KnowIT VQA, fusing video, \sceneSum and \episodeSum input sources. All fusion methods use soft temporal attention for localization of episode input sources. Top: baseline/competitors. Bottom: ours.}
\label{tab:fusion}
\end{table}
\paragraph{Multi-\branch results} 

We evaluate our two multi-\branch QA methods introduced in \autoref{sec:fusion}, namely \emph{multi-\branch attention} and \emph{self-attention}, comparing them with the following combinations/baselines/competitors:\\

\begin{enumerate}[itemsep=3pt, parsep=0pt, topsep=2pt]
	\item \emph{Multi-\branch self-attention}: combination of multi-\branch attention and self-attention: the output of the latter is weighted by the former. The remaining pipeline is the same as in multi-\branch attention.
	\item \emph{Product}: Hadamard product on embeddings of all \branches per answer, followed by a linear classifier per answer. The remaining pipeline is the same.
	\item \emph{Modality weighting}~\cite{garcia2020knowledge}: a linear classifier~\eq{short-class} and loss function is used as in single-\branch QA but with transformers frozen for each \branch separately. The obtained scores by single-\branch classifiers are combined by a multi-\branch classifier and another loss function applies. The overall loss all is a linear combination with weight $\beta_\omega$ on the multi-\branch loss and $1-\beta_\omega$ uniformly distributed over single-\branch losses.
\end{enumerate}

\autoref{tab:fusion} shows results for fusion of video, \sceneSum and \episodeSum. For modality weighting, we set $\beta_\omega = 0.7$ according to the validation set. Our multi-\branch attention outperforms other fusion methods. Besides, it does not require tuning of modality weight hyperparameter $\beta_\omega$ or selecting the number of heads and blocks for self-attention. Unless specified, we use multi-stream attention for fusion by default.

\begin{table}
\centering
\small
\setlength{\tabcolsep}{2pt}
\begin{tabular}{lccccc} \toprule
\Th{Method}                                & \Th{Vis.}  & \Th{Text.} & \Th{Temp.} & \Th{Know.} & \best{\Th{All}}   \\ \midrule
ROLL~\cite{garcia2020knowledge}$\dagger$   & 0.722      & 0.703      & 0.709      & 0.697      & 0.704             \\
$+$ Multi-\branch attention                & 0.724 & 0.721      & 0.721      & 0.691      & 0.703             \\
$+$ More parts for plot                    & 0.722      & 0.703      & 0.651      & 0.717      & 0.714             \\
$+$ New order of plot inputs               & 0.730      & 0.710      & 0.686      & 0.712      & 0.715             \\
$+$ Temporal attention                     & 0.734      & 0.725      & 0.663      & 0.724      & 0.724             \\
$\pm$ Replacing P $\rightarrow$ E            & 0.753      & \tb{0.815}      & \tb{0.814}      & \tb{0.773}    & 0.775
\\ 
$\pm$ Replacing D $\rightarrow$ S            & \tb{0.755} & 0.783 & 0.779 & 0.789      & \best{\tb{0.781}}             \\ \bottomrule
\end{tabular}%
\caption{\emph{Accuracy improvements over} ROLL~\cite{garcia2020knowledge}. $\dagger$: our reproduction. Each row adds a new improvement except the last two, where we replace streams. P: plot; E: \episodeSum; D: dialog; S: \sceneSum.}
\label{tab:roll-improve}
\end{table}

\paragraph{Improvements over~\cite{garcia2020knowledge}}

We reproduce ROLL~\cite{garcia2020knowledge} using official code by the authors and default parameters. This is our baseline, shown in the first row of \autoref{tab:roll-improve}. Then, we evaluate our improvements, adding them one at a time. First, we replace modality weighting with \emph{multi-\branch attention}. Despite its simplicity, its performance is on par, losing only 0.1\%, while requiring no hyperparameter tuning. Then, we increase the \emph{number of parts} of plot summaries from 5 to 10, eliminating information loss by truncation and bringing an accuracy improvement of 1.1\%. We change the \emph{order of arguments} of BERT for episode input sources from $f(q, a^c + p_j)$ to $f(p_j + q, a^c)$~\eq{long-bert}, which is consistent with~\eq{short-bert} and improves only slightly by 0.1\%. Our new \emph{temporal attention} mechanism improves accuracy by 0.9\%. Replacing plot with \episodeSum, which is our main contribution, brings an improvement of 5.1\%. Finally, the accuracy is improved by 0.6\% by using \emph{\sceneSum} instead of raw dialog. The overall gain over~\cite{garcia2020knowledge} is 7.7\%.

Note that the relative improvement of each new idea depends on the order chosen in \autoref{tab:roll-improve}. For instance, the order of BERT arguments brings improvements of up to 2.3\% in experiments including the \episodeSum.

\section{Conclusion}
\label{sec:conclusion}

KnowIT VQA is a challenging dataset where it was previously believed that some form of external knowledge was needed to handle knowledge questions, as if knowledge was yet another modality. Our results indicate that much of this required knowledge was hiding in \emph{dialog}, waiting to be harnessed. It is also interesting that our \emph{soft temporal attention} helps a lot more with our episode dialog summary than human plot summary, which may be due to the episode dialog summary being longer. This may also explain the astounding performance of episode dialog summary, despite its low overall quality: plot summaries are of much higher quality but may be missing a lot of information.


\paragraph{Acknowledgements}
This work was supported by the European Commission under European Horizon 2020 Programme, grant number 951911 - AI4Media. This work was granted access to the HPC resources of IDRIS under the allocation 2020-AD011012263 made by GENCI.

\clearpage

{\small
\balance
\bibliographystyle{ieee_fullname}
\bibliography{egbib}
}

\clearpage
\nobalance

\appendix

\section*{Appendix}
\section{Additional qualitative analysis}
\label{sec:add-qual}

\paragraph{Dialog summarization}

\begin{figure}[ht!]
\centering
\fig{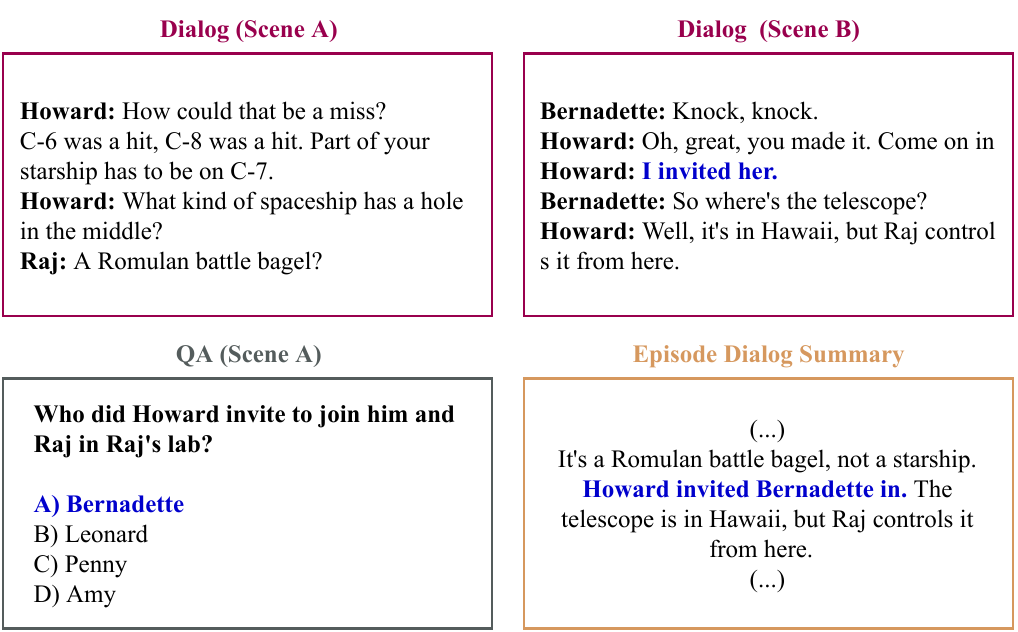}
\caption{Dialog summarization converts pronouns in dialog to character names in \episodeSum, supporting question answering. In particular, ``I'' is substituted by ``Howard'' and ``her'' by ``Bernadette''.}
\label{fig:pronoun}
\end{figure}

In the example of \autoref{fig:pronoun}, Howard says, ``I invited \emph{her}.'' in scene B. Our dialog summarization interprets this sentence by assigning the correct character name: ``Howard invited \emph{Bernadette} in.'' Hence, we can answer the question of scene A, ``Who did Howard invite to join him and Raj in Raj's lab?'' correctly. Thanks to the episode dialog summary spanning all scenes and the use of character names instead of pronouns, our method can answer character-related questions correctly.

\paragraph{Plot \vs \episodeSum}

\begin{figure}[h]
\centering
	\fig[.95]{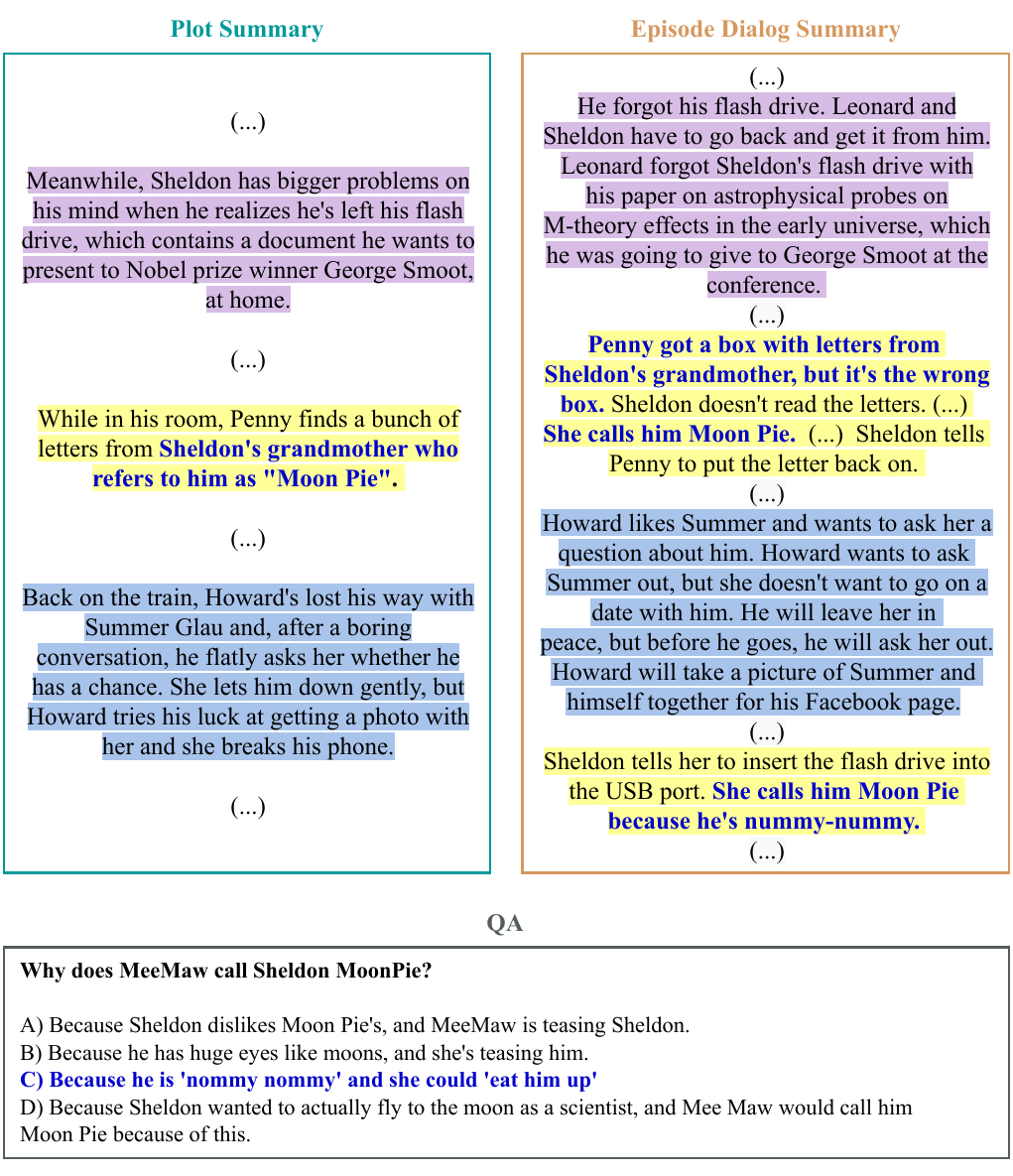}
\caption{An example of plot summary and \episodeSum, with each topic highlighted in the same color in both summaries. Phrases relevant to QA in blue. Only the \episodeSum contains enough information to answer the question.}
\label{fig:plot_episode_summary}
\end{figure}

A comparison of plot summary and \episodeSum is given in \autoref{fig:plot_episode_summary}. There are three different topics in the story line, and each is highlighted with the same color in both summaries. The first topic, highlighted in purple, is ``Sheldon's forgotten flash drive.''  The second, highlighted in yellow, is ``Sheldon's grandmother.'' The third, highlighted in blue, is ``Asking Summer out.'' The plot summary is topic-centered, while the \episodeSum is following the narrative order. Hence, topics may be fragmented in the latter. The \episodeSum has more detail than the plot. In particular, it contains enough information to answer the question \emph{Why does Sheldon's grandmother call him Moon Pie?} That is, \emph{because he's nummy-nummy}. This information is missing from the plot summary, which focuses on the main topics/events of an episode. Even though the \episodeSum is noisy, it contains details that help in question answering.


\begin{figure*}[ht]
\centering
\small
\begin{tabular}{cc}
	\fig[.48]{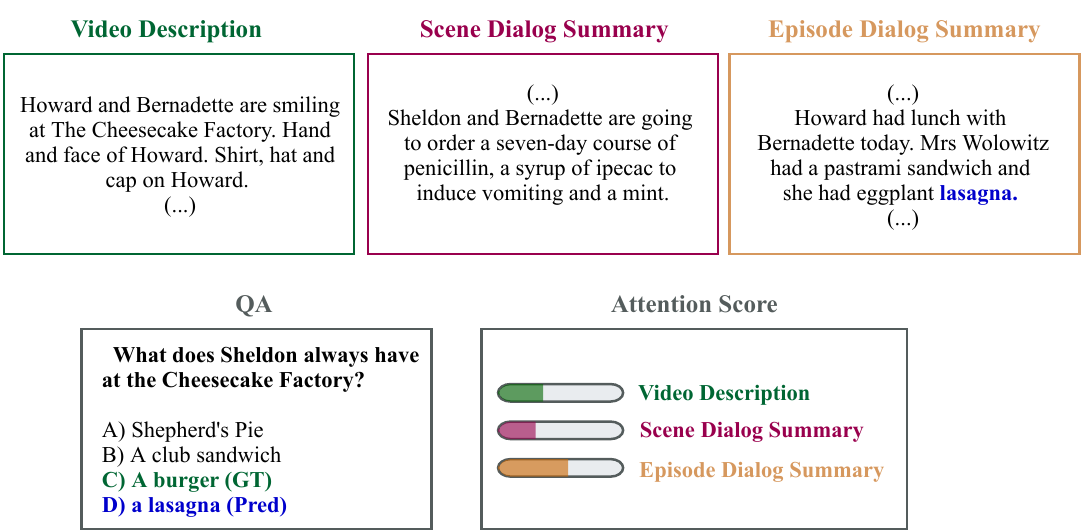} &
	\fig[.48]{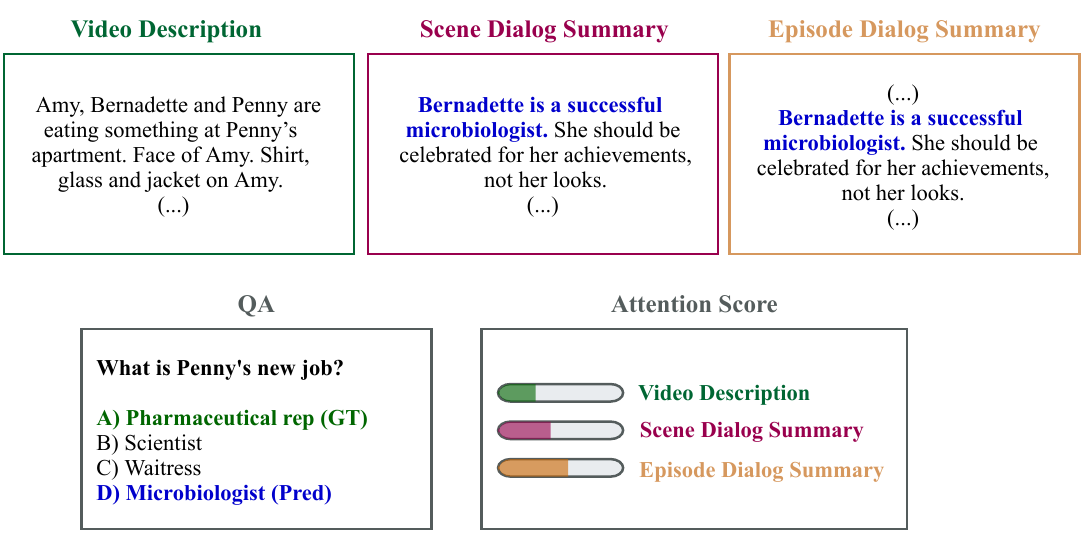} \\
	(a) Knowledge QA (1) &
	(b) Textual QA (1)\\
	\fig[.48]{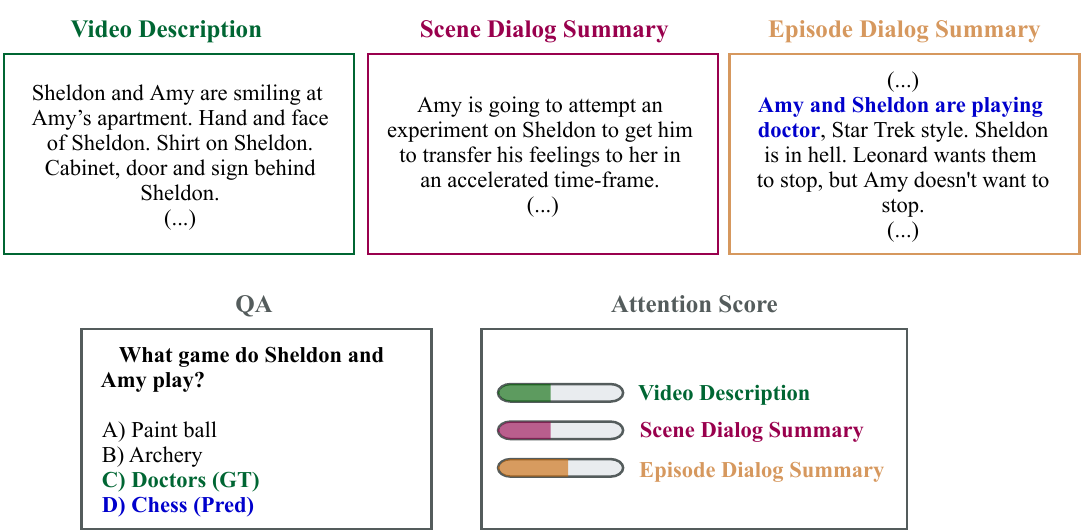} &
	\fig[.48]{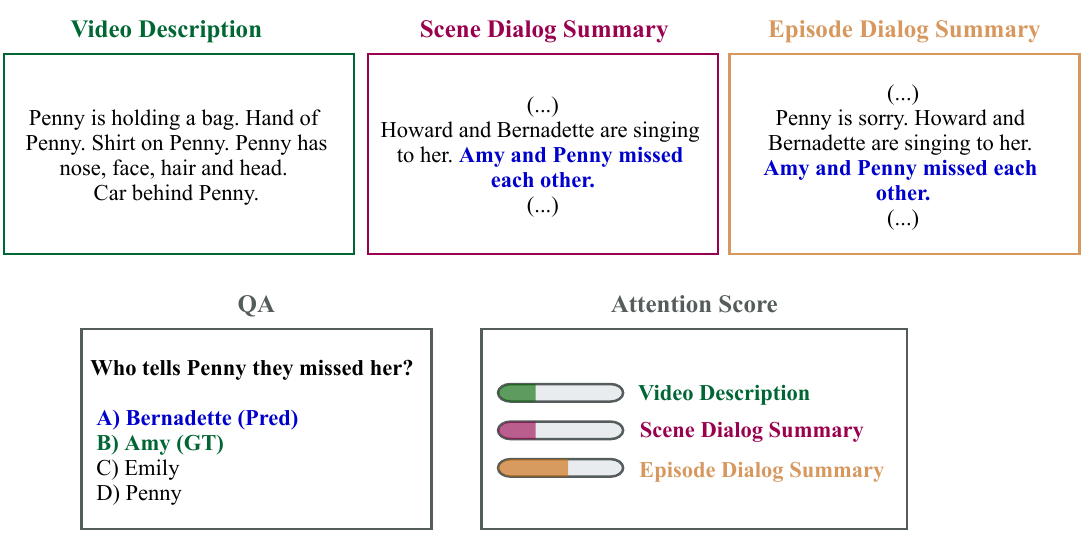} \\
	(c) Knowledge QA (2)&
	(d) Textual QA (2)\\
	\fig[.48]{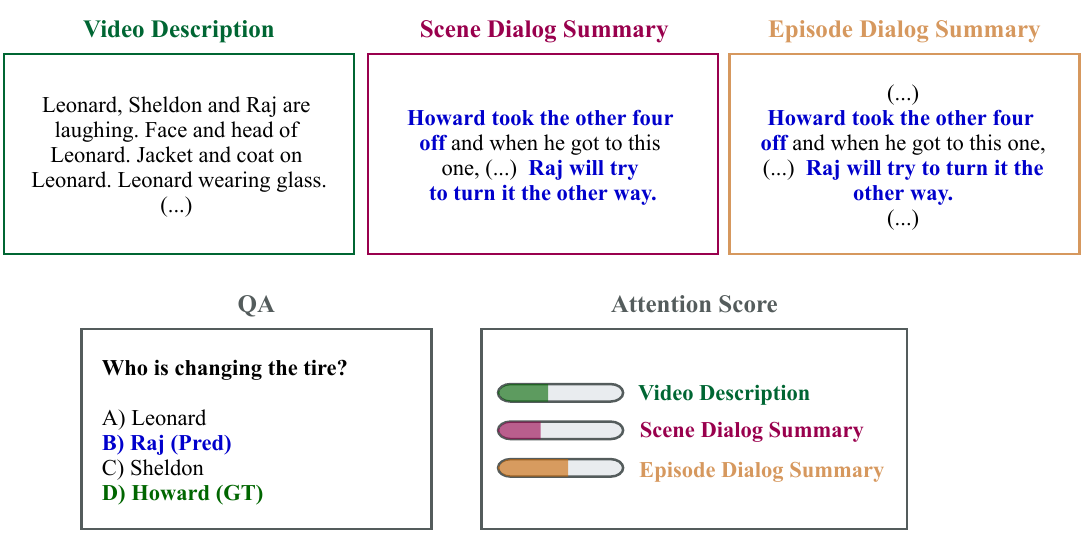} &
	\fig[.48]{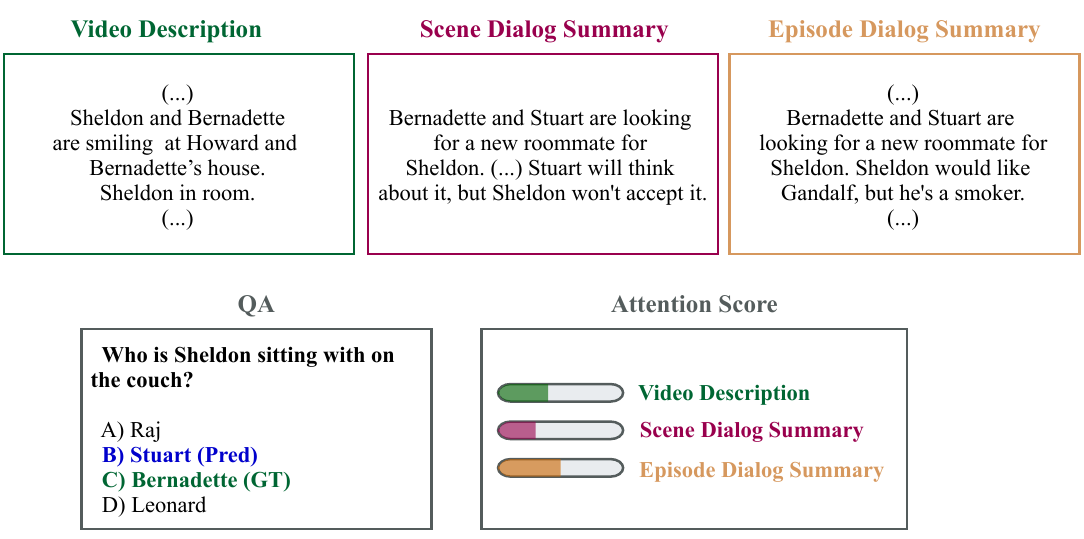} \\
	(e) Temporal QA &
	(f) Visual QA
\end{tabular}
\caption{\emph{Failed predictions of multi-\branch attention}. We highlight in blue the part of the source text that might be relevant to answering the question. ``Pred''/blue: model predictions. ``GT''/green: ground truth.}
\label{fig:qual_failure}
\end{figure*}

\paragraph{Failure cases}

\autoref{fig:qual_failure} shows examples of failed predictions of our model along with \branch attention scores for different question types. The model receives three input sources, question/answers and attention scores over inputs.

\autoref{fig:qual_failure}(a) refers to a knowledge question, which requires recurrent knowledge of the whole TV show. In other words, the correct answer cannot be found in \episodeSum. The question is answered as "a lasagna" found in \episodeSum, even though it is wrong.

\autoref{fig:qual_failure}(b) refers to a textual question, which should have been answered by \sceneSum. However, \sceneSum does not contain the correct answer. Our model gives most attention to \episodeSum. The prediction is made according to the highlighted text, which is the same in both sources. However, this prediction refers to the wrong person.

\autoref{fig:qual_failure}(c) refers to another knowledge question, which could be answered by the highlighted text in \episodeSum. Even though \episodeSum has the most attention, the prediction is incorrect.

\autoref{fig:qual_failure}(d) refers to another textual question, which should have been answered by \sceneSum. Although both \sceneSum and \episodeSum include the correct answer, and \episodeSum has the most attention, the prediction indicates the wrong person.\\

\autoref{fig:qual_failure}(e) refers to a temporal question. The \sceneSum and \episodeSum imply that \emph{Raj} and \emph{Howard} might be changing the tire. The video description is not helpful either. Hence, our model predicts \emph{Raj}, while the correct answer is \emph{Howard}.

\autoref{fig:qual_failure}(f) is a visual question. However, the video description fails to convey relevant information to answer the question. The other inputs do not contain relevant information either. One of the character names appearing in \episodeSum is predicted, which is incorrect.

\section{Additional ablation studies}
\label{sec:add-ablation}

\paragraph{Hyperparameter validation}

\begin{figure}
\centering
\small
\begin{tikzpicture}
\begin{axis}[
	height=.6\columnwidth,
	xlabel={$\beta_\omega$},
	ylabel={validation accuracy},
	y tick label style={
    /pgf/number format/.cd,
    fixed,
    fixed zerofill,
    precision=3
  },
   legend style={font=\tiny},
   legend pos= south east,
]
\pgfplotstableread{
	beta acc_soft acc_hard
	0.1  0.739 0.746
	0.2  0.749 0.760
	0.3  0.751 0.761
	0.4  0.750 0.761
	0.5  0.749 0.760
	0.6  0.750 0.761
	0.7  0.752 0.761
	0.8  0.750 0.760
	0.9  0.747 0.755
	1.0  0.756 0.751
}{\acc}
	\addplot table[x=beta,y=acc_soft]  {\acc};
	\addlegendentry{Soft Attn.}
	\addplot table[x=beta,y=acc_hard] {\acc};
    \addlegendentry{Hard Attn.}
\end{axis}
\end{tikzpicture}
\caption{Accuracy \vs $\beta_\omega$ for fusion of video, \sceneSum and \episodeSum by modality weighting~\cite{garcia2020knowledge} on KnowIT VQA validation set.}
\label{fig:beta_selection}
\end{figure}
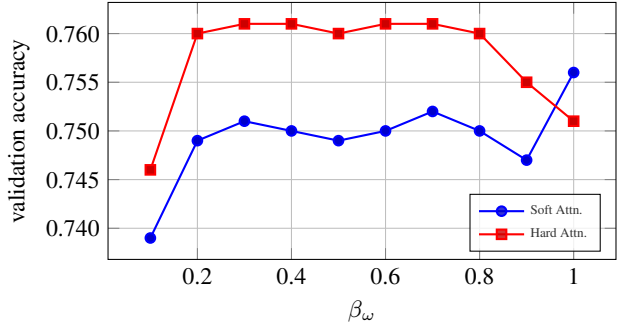

\emph{Modality weighting}~\cite{garcia2020knowledge} fusion method requires selection of hyperparameter $\beta_{\omega}$. \autoref{fig:beta_selection} shows validation accuracy \vs $\beta_\omega$ for fusion of video, \sceneSum and \episodeSum. We choose $\beta_{\omega} = 0.7$ for both soft and hard temporal selection to report results in \autoref{tab:fusion} and \autoref{tab:softmax_fusion}. The remaining weight of $1-\beta_{\omega}$ is evenly distributed over individual \branch losses as $0.1$ per \branch.

\paragraph{Effect of temporal attention on single-\branch QA}

We investigate the effect of our soft temporal attention (\autoref{sec:long}) on single-stream QA for episode input sources. We also evaluate the effect of single-stream training with soft or hard temporal attention on multi-stream attention, where we use soft temporal attention. According to \autoref{tab:softmax}, temporal attention improves the accuracy of plot and \episodeSum by 1.9\% and 3.3\%, respectively. Accordingly, the accuracy of multi-\branch QA on the same episode sources as well as video and \sceneSum increases by 0.6\% and 7.0\%, respectively. The gain is higher when \episodeSum is used, since the \episodeSum is longer than plot.

\begin{table}
\centering
\small
\setlength{\tabcolsep}{1.6pt}
\begin{tabular}{llcccccc} \midrule
\mr{2}{\Th{Stream}}     & \mr{2}{\Th{Inputs}}  & \Th{Soft}  & \mr{2}{\Th{Vis.}} & \mr{2}{\Th{Text.}} & \mr{2}{\Th{Temp.}} & \mr{2}{\Th{Know.}} & \mr{2}{\best{\Th{All}}} \\ 
                        &                      & \Th{Attn.}              \\ \midrule
\multirow{4}{*}{Single} & P                    & -           & 0.656             & 0.594              & 0.628              & 0.712              & 0.683                   \\
                        & P                    & \ch         & 0.666             & 0.623              & 0.593              & 0.735              & 0.702                   \\
                        & E                    & -           & 0.604             & 0.721              & 0.733         & 0.765              & 0.723                   \\
                        & E                    & \ch         & \tb{0.676}        & \tb{0.750}         & \tb{0.779}              & \tb{0.785}         & \best{\tb{0.756}}       \\ \midrule
\multirow{4}{*}{Multi}  & V + S + P            & -           & 0.732             & 0.688              & 0.674              & 0.720              & 0.717                   \\
                        & V + S + P            & \ch         & 0.739             & 0.699              & 0.628              & 0.728              & 0.723                   \\
                        & V + S + E            & -           & 0.707        & 0.772        & 0.721              & 0.700              & 0.711                   \\
                        & V + S + E            & \ch         & \tb{0.755}             & \tb{0.783}              & \tb{0.779}              & \tb{0.789}         & \best{\tb{0.781}}       \\ \midrule
\end{tabular}
\caption{
\emph{Effect of temporal attention on single-\branch QA} on KnowIT VQA. Soft Attn.: soft temporal attention on single-stream training. We use soft temporal attention for multi-\branch QA, but this is still affected by the temporal attention used in single-stream training. V: video; S: \sceneSum; P: plot; E: \episodeSum.}
\label{tab:softmax}
\end{table}


\paragraph{Effect of temporal attention on multi-\branch QA}

\autoref{tab:softmax_fusion} shows the effect of \emph{soft temporal attention} on multi-\branch QA for fusion of video, \sceneSum and \episodeSum input sources. We use soft temporal attention for single-\branch QA of \episodeSum. In all fusion methods, the overall accuracy is improved by using soft temporal attention.\\

\begin{table}
\centering
\small
\setlength{\tabcolsep}{1.6pt}
\begin{tabular}{lcccccc} \midrule
\mr{2}{\Th{Method}}      & \Th{Soft}  & \mr{2}{\Th{Vis.}} & \mr{2}{\Th{Text.}} & \mr{2}{\Th{Temp.}} & \mr{2}{\Th{Know.}} & \mr{2}{\best{\Th{All}}} \\ 
                        &       \Th{Attn.}           \\ \midrule
\multirow{2}{*}{Product} & -           & 0.728             & 0.645              & 0.744              & 0.756              & 0.736                   \\
                        & \ch         & 0.743             & 0.659              & 0.756              & 0.751              & 0.739                   \\ \midrule
Modality weighting       & -           & 0.716              & \tb{0.815}              & 0.791              & 0.776              & 0.768                   \\
\cite{garcia2020knowledge}  & \ch         & 0.708             & 0.786              & 0.767              & 0.787              & 0.769                   \\ \midrule
\multirow{2}{*}{Self-attention} & -           & 0.753             & 0.804              & \tb{0.802}              & 0.766              & 0.769                   \\
                        & \ch         &\tb{0.759}            & 0.764              & 0.767              & 0.777              & 0.771                   \\ \midrule
\multirow{2}{*}{Multi-steam attention} & -           & 0.743             & 0.790              & 0.779              & 0.785              & 0.776                   \\
                        & \ch         & 0.755             & 0.783              & 0.779              & \tb{0.789}              & \tb{0.781}                   \\ \midrule
\multirow{2}{*}{Multi-steam self attn.} & -           & 0.749             & 0.797              & 0.791              & 0.768              & 0.768                   \\
                        & \ch         & 0.755             & 0.768              & 0.756              & 0.777              & 0.770                   \\ \midrule

\end{tabular}
\caption{\emph{Effect of temporal attention on multi-\branch QA} on KnowIT VQA for fusion of video, \sceneSum, and \episodeSum  input sources. Soft Attn.: soft temporal attention on multi-stream training. We use soft temporal attention for single-stream QA of \episodeSum.}
\label{tab:softmax_fusion}
\end{table}

\begin{table}[ht]
\centering
\small
\setlength{\tabcolsep}{2pt}
\newcommand{\dvp}{\begin{tabular}[c]{@{}l@{}}D \\ V \\ P \end{tabular}}
\newcommand{\dvps}{\begin{tabular}[c]{@{}l@{}}D \\ V \\ P \\ S \end{tabular}}
\newcommand{\dvpse}{\begin{tabular}[c]{@{}l@{}}D \\ V \\ P \\ S\\ E \end{tabular}}
\begin{tabular}{cllllll} \toprule
\Th{Analysed}   & \mr{2}{\Th{Inputs}} & \mr{2}{\Th{Vis.}} & \mr{2}{\Th{Text.}} & \mr{2}{\Th{Temp.}} & \mr{2}{\Th{Know.}} & \mr{2}{\best{\Th{All}}} \\ 
\Th{Inputs}                                                                                                                                        \\ \midrule
\mr{3}{\dvp}    & D+V                 & 0.693             & 0.768              & 0.593              & 0.554              & 0.611                   \\
                & D+P                 & 0.732             & 0.721              & 0.674              & 0.723              & 0.723                   \\ \cmidrule{2-7} 
                & D+V+P               & 0.734             & 0.725              & 0.663              & 0.724              & 0.724                   \\ \midrule
\mr{4}{\dvps}   & D+S                 & 0.664             & 0.786         & 0.628              & 0.548              & 0.604                   \\
                & V+S                 & 0.689             & 0.721              & 0.581              & 0.549              & 0.601                   \\
                & P+S                 & 0.716             & 0.710              & 0.628              & 0.727              & 0.719                   \\ \cmidrule{2-7} 
                & D+V+P+S             & 0.734        & 0.732              & 0.663              & 0.725              & 0.726                   \\ \midrule
\mr{7}{\dvpse} & D+E                 & 0.743             & 0.812              & 0.779              & 0.779              & 0.775                   \\
                & V+E                 & 0.732             & 0.761              & 0.767         & 0.788              & 0.772                   \\
                & P+E                 & 0.716             & 0.743              & 0.721              & \tb{0.791}              & 0.766                   \\
                \cmidrule{2-7} 
                & D+S+E               & 0.743             & \tb{0.822}              & \tb{0.802}         & 0.771              & 0.772                   \\
                & V+S+E               & \tb{0.755}             & 0.783              & 0.779         & 0.789              & \best{\tb{0.781}}                   \\
                 & P+S+E               & 0.739             & 0.779              & 0.733    
                & 0.783              & 0.771                   \\
                \cmidrule{2-7} 
                & D+V+P+S+E           & 0.751             & 0.797              & 0.744              & 0.781              & 0.775                   \\ \bottomrule
\end{tabular}
\caption{\emph{Multi-\branch QA accuracy} on KnowIT VQA: comparison of different input combinations for multi-\branch attention. D: dialog; V: video; P: plot; S: \sceneSum; E: \episodeSum.}
\label{tab:fusion_results_soft}
\end{table}

\paragraph{Different input combinations}

\autoref{tab:fusion_results_soft} shows the accuracy of multi-stream QA for different input combinations, where the number of input \branches~varies in $\{2, 3, 4, 5\}$. Scene dialog summaries improves the accuracy compared with single-stream QA results in \autoref{tab:branch_results}. Moreover, using the \episodeSum always improves the overall accuracy by a large margin. The best overall accuracy of 0.781 is achieved by video, \sceneSum, and \episodeSum.


\paragraph{Question type $\leftrightarrow$ attention scores}

We perform significance testing for the dependence between the question type and attention scores. There are 2 independent variables in the scores of 3 streams, whose values we discretize into $10 \times 10$ bins. We form a $4 \times 10 \times 10$ joint histogram of question type ($X$) and scores ($Y$) and compute the mutual information $I(X;Y)$. We perform a $G$-test\footnote{https://en.wikipedia.org/wiki/G-test} with $G = 2N \cdot I(X;Y)$, where $N=2361$ is the number of test questions. Finally, using a chi-square distribution of $3 \times 9 \times 9$ DoF, we find a $p$-value of $1.52 \times 10^{-25}$ for the null hypothesis. This indicates that attention scores depend on question type.


\paragraph{Replacing attention scores with oracle scores determined by question type}

Assuming that we know the question type for the test set, we perform an \emph{oracle} experiment where attention scores are based on question type rather than our fusion method. We only consider visual, textual, and knowledge types of question. In particular, we assign visual questions to video input, textual questions to \sceneSum and knowledge questions to \episodeSum. We exclude temporal questions since they can be answerable by \sceneSum or video. Only 3.6\% of questions are of temporal type in the test set. We find that our multi-stream attention method (0.781\%) is 3.6\% better than the oracle experiment (0.745\%). This indicates that our fusion mechanism is more effective than a na\"ive oracle that assumes more knowledge.

\end{document}